%% file: main.tex
\documentclass[final,1p,times]{elsarticle}
\usepackage{xcolor}
\usepackage{lineno}
\usepackage{booktabs}
\usepackage{subcaption}
\usepackage{url}
\newcommand{\newtext}[1]{\textcolor{black}{#1}}

\input{preamble}

\usepackage{amssymb}

\usepackage{amsmath}
\usepackage{cleveref}
\usepackage{algorithmic}
\usepackage[ruled]{algorithm2e}

\usepackage{soul}

\crefname{figure}{Fig.}{Figs.}
\crefname{appendix}{}{}

\begin{document}

\begin{frontmatter}

\author[tue]{Tim J. Schoonbeek\fnref{fn1}\corref{cor1}}
\author[tue]{Shao-Hsuan Hung\corref{cor1}}
\author[tue]{Dan Lehman}
\author[asml]{Hans Onvlee}
\author[asml]{Jacek Kustra}
\author[tue]{Peter H.N. de With}
\author[tue]{Fons van der Sommen}

\title{Learning to Recognize Correctly Completed Procedure Steps in Egocentric Assembly Videos through Spatio-Temporal Modeling\tnoteref{aam}}

\affiliation[tue]{organization={Dept. of Electrical Engineering, Eindhoven University of Technology}, addressline={P.O. Box 513}, postcode={5600 MB}, city={Eindhoven}, country={Netherlands}}
\affiliation[asml]{organization={ASML}, addressline={De Run 6501}, postcode={5504 DR}, city={Veldhoven}, country={Netherlands}}

\cortext[cor1]{equal contribution}
\fntext[fn1]{Corresponding author: Tim J. Schoonbeek, email: t.j.schoonbeek@tue.nl}
\tnotetext[aam]{\textit{Author accepted manuscript (AAM)}.
Accepted for publication in \textit{Computer Vision and Image Understanding}.
The Version of Record will be available in due course; this arXiv record will be updated with the DOI once assigned.
Please cite the published article when available.
This AAM is made available under the CC BY-NC-ND 4.0 license.}

\input{0_abstract}


\end{frontmatter}

\input{1_intro}
\input{2_related}
\input{3_method}
\input{4_experiment}
\input{5_discussion}

\input{6_conclusion}

{
    \small
    \bibliographystyle{ieeenat_fullname}
    \bibliography{main}
}
\newpage 
\input{X_suppl}

\end{document}

%% file: preamble.tex
\usepackage{soul}

\usepackage{pifont}
\newcommand{\cm}{\ding{51}}

\DeclareRobustCommand\onedot{\futurelet\@let@token\@onedot}
\def\@onedot{\ifx\@let@token.\else.\null\fi\xspace}

\def\eg{\emph{e.g.}} 

\def\ie{\emph{i.e. }}

\def\etal{\emph{et al. }}

%% file: 0_abstract.tex
\begin{abstract}
Procedure step recognition (PSR) aims to identify all correctly completed steps and their sequential order in videos of procedural tasks. 
The existing state-of-the-art models rely solely on detecting assembly object states in individual video frames.
By neglecting temporal features, model robustness and accuracy are limited, especially when objects are partially occluded.
To overcome these limitations, we propose Spatio-Temporal Occlusion-Resilient Modeling for Procedure Step Recognition (STORM-PSR), a dual-stream framework for PSR that leverages both spatial and temporal features.
The assembly state detection stream operates effectively with unobstructed views of the object, while the spatio-temporal stream captures both spatial and temporal features to recognize step completions even under partial occlusion.
This stream includes a spatial encoder, pre-trained using a novel weakly supervised approach to capture meaningful spatial representations, and a transformer-based temporal encoder that learns how these spatial features relate over time.
STORM-PSR is evaluated on the MECCANO and IndustReal datasets, reducing the average delay between actual and predicted assembly step completions by 11.2\% and 26.1\%, respectively, compared to prior methods.
We demonstrate that this reduction in delay is driven by the spatio-temporal stream, which does not rely on unobstructed views of the object to infer completed steps.
The code for STORM-PSR, along with the newly annotated MECCANO labels, is made publicly available at \url{https://timschoonbeek.github.io/stormpsr}.
\end{abstract}

%% file: 1_intro.tex
\section{Introduction}
\label{sec:intro}
In the field of assistive computer vision in industrial contexts, understanding and tracking key steps in videos of procedure execution is an active line of research~\cite{Assembly101,grauman2024ego,EPIC-KITCHEN,lin2022learning,zhong2023learning,nagasinghe2024not,ashutosh2024video}. 
Procedural tasks often require multiple interdependent actions to be executed correctly, although reliably recognizing these steps remains challenging due to variable viewing angles, frequent occlusions, and flexible execution orders. 
Particularly egocentric cameras introduce dynamic perspectives and frequent hand occlusions, obscuring key objects or actions and creating gaps in visual continuity that complicate accurate procedural tracking~\cite{MECCANO, IndustReal, grauman2022ego4d}. Such egocentric technologies can be beneficial for assistive technology in industrial scenarios~\cite{surveyforegocentric,CV4assist,DL4assistiveCV}.

\begin{figure}
    \centering
    \includegraphics[width=0.75\linewidth]{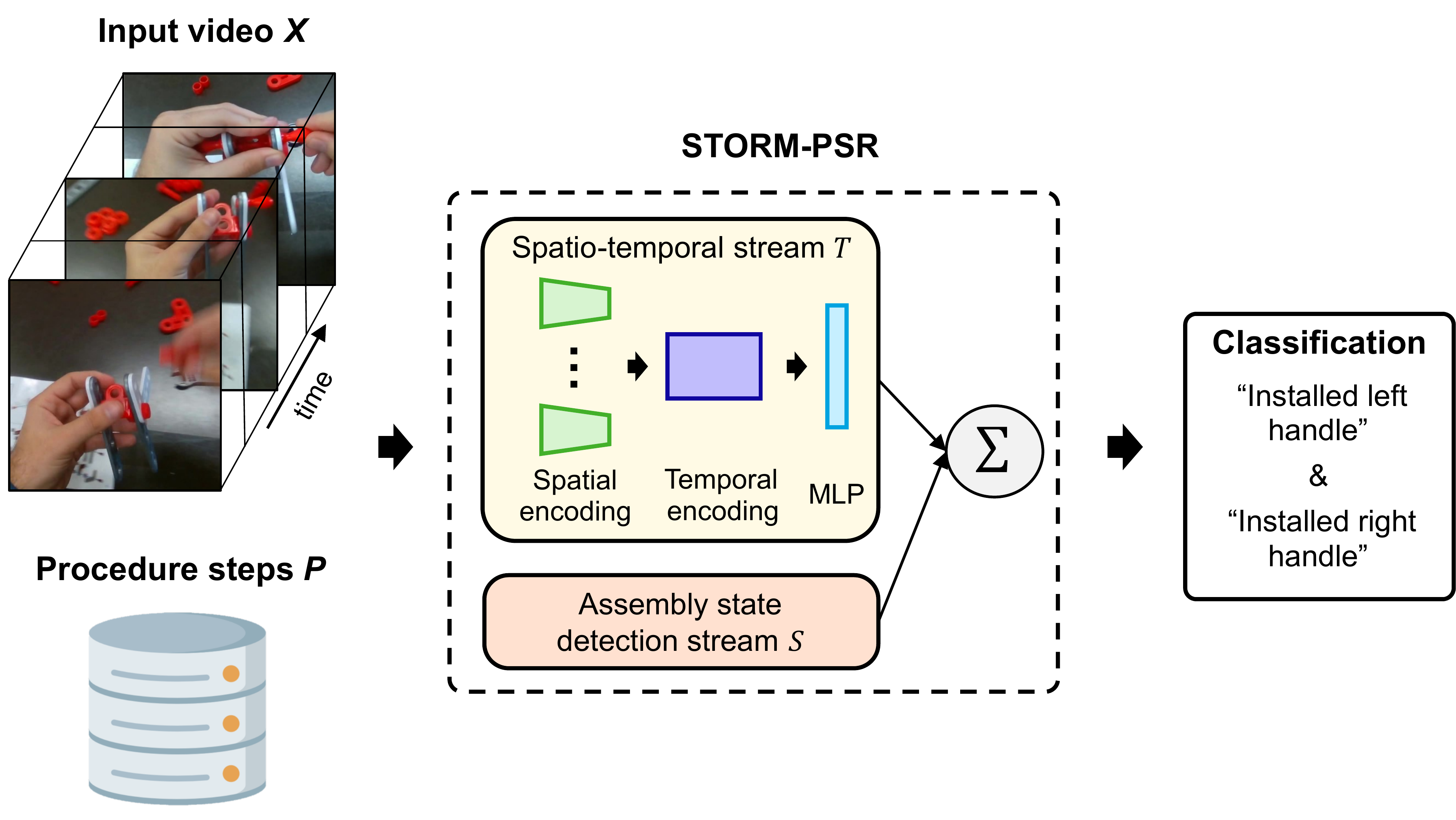}
    \caption{Diagram of the proposed Spatio-Temporal Occlusion-Resilient Modeling for Procedure Step Recognition (STORM-PSR) approach, using assembly state detection for non-occluded frames and temporal modeling to recognize step completions when the assembly object is only partially visible.} 
    \label{fig: teaser}
    \vspace{-0.2cm}
\end{figure}

Existing work on recognizing and tracking procedural videos take multiple approaches, such as action recognition~\cite{grauman2024ego,Assembly101,MECCANO,EPIC-KITCHEN}, procedural activity understanding~\cite{lin2022learning,zhong2023learning,nagasinghe2024not}, keystep recognition~\cite{ashutosh2024video,shah2023steps,grauman2024ego,bansal2022my}, \newtext{procedural error identification~\cite{prego}}, and procedure step recognition~\cite{IndustReal}.
In procedural action recognition, the objective is to categorize actions performed by humans in short video clips.
In contrast, the task of procedural \emph{activity} recognition focuses on understanding long-form videos. 
Within this, the task of \emph{keystep} recognition aims to find key moments and their sequences in long videos of procedure executions.
Crucially, the aforementioned approaches lack a measure of completeness and correctness for the recognized actions and steps.
Schoonbeek~\etal~\cite{IndustReal} address this limitation by proposing a \emph{procedure step recognition} (PSR) task, aiming to identify all \textit{correctly completed} procedural steps in long-form videos, each contributing to the successful execution of a procedure. 
However, their approach is fundamentally limited by its reliance on assembly state detections (ASD)~\cite{ASD-1,ASD-2,ASD-3,ASDF,origami}, which only provide reliable predictions for unobstructed views of the object.
\newtext{Such unobstructed video frames are rare, particularly in egocentric videos, where the presence of hands and tools frequently occlude the object of interest due to the nature of hand-object interaction during task execution.
Moreover, ASD is based exclusively on spatial information, operating on single frames. 
Contrarily, PSR is a temporal video understanding task, requiring reasoning over video sequences to assess which steps have been successfully completed. 
Yet, none of the existing PSR methods have explicitly leveraged temporal features for this purpose. 
Additionally, the existing approaches are only evaluated on a single dataset, namely the IndustReal~\cite{IndustReal} dataset. 
Therefore, the generalization to other data remains unknown.
This work explores the hypothesis that procedural step recognition does not require full visibility of the object, but only of the components pertinent to the current step completion.
The aim is to determine whether a spatio-temporal model, trained directly on PSR labels, can address the constraints imposed by ASD-based methods.}
Specifically, STORM-PSR, outlined in \cref{fig: teaser}, is introduced. STORM-PSR is a two-stream approach for PSR that combines ASD with a spatio-temporal model to recognize procedure steps even under partial occlusions.

The ASD stream detects the assembly stage in each video frame and provides a measure of correctness, since erroneous states can be defined and detected.
The PSR task can be addressed through ASD, by inferring which steps are correctly executed between observations of assembly states using prior knowledge of the assembly procedure~\cite{IndustReal}. 
Whenever unobstructed images are available, the robust ASD algorithm provides reliable predictions. 
However, such unobstructed frames are rare in egocentric videos due to hand occlusions and camera motion, resulting in significant delays between the actual completion of a step and its subsequent recognition.
Therefore, we propose a second stream, comprising a transformer-based temporal encoder to aggregate spatial features even when only obstructed images are available.
This so-called spatio-temporal stream directly predicts correctly completed procedure steps, rather than inferring them between consecutive observations of assembly states.
It therefore only requires visibility of the parts of the object that constitute a specific step.
For example, to recognize the completion of the step `install front wheel', the visibility of irrelevant parts, \eg the rear wheel, is not required.

These direct PSR predictions are facilitated through a multi-layer perceptron (MLP) head following the temporal transformer, performing multi-label classification to handle multiple step completions within a single video clip.
We introduce key-frame sampling (KFS), a weakly supervised pre-training approach that leverages sparse step-completion annotations.
KFS enables the spatial encoder to provide meaningful representations of assembly states, using a contrastive loss that maximizes agreement between the same states from different videos and occlusion levels, while minimizing the agreement with different states.
When available, synthetic data of assembly states can be readily integrated with KFS, thereby reducing reliance on annotated real-world data.
For training the temporal transformer, we propose key-clip aware sampling (KCAS).
KCAS is based on a bimodal distribution to over-sample step completions (positives) and video frames that are shortly preceding these step completions (hard negatives).
Simultaneously, KCAS under-samples potentially ambiguous frames surrounding the exact moment of completion and less informative clips that contain no step completions.
Trained with KFS and KCAS, STORM-PSR reduces the average delay ($\tau$) between the completion and corresponding recognition of a step by 26.1\% on the IndustReal dataset~\cite{IndustReal} and 11.2\% on the MECCANO dataset~\cite{MECCANO}, which we have annotated with PSR and ASD labels to establish the performance benchmark for this task.
In summary, this paper contains the following contributions.
\begin{itemize}
    \item STORM-PSR: a dual-stream approach that leverages temporal features for PSR, thereby significantly reducing the average delay compared to state-of-the-art methods.
    \item Key-frame sampling (KFS) for weakly supervised pre-training of the spatial encoder, and key-clip aware sampling (KCAS), a novel sampling strategy for training the temporal encoder.
    \item Annotations for the MECCANO dataset~\cite{MECCANO} and a corresponding performance benchmark for PSR and ASD.
\end{itemize}

\newtext{The remainder of this article is structured as follows. \Cref{sec:related work} reviews prior work relevant to procedure step recognition, after which \Cref{sec:psr} introduces the formal definition of the task and the evaluation metrics used. The proposed framework, STORM‑PSR, is described in detail in \Cref{sec:method}. \Cref{sec:exp} reports experimental results on the IndustReal and MECCANO datasets. \Cref{sec:discussion} examines the implications of the findings and outlines key limitations. Finally, \Cref{sec:conclusion} summarizes the main contributions.}

%% file: 2_related.tex
\section{Related work}
\label{sec:related work}

\subsection{Procedure understanding}
\label{sec:rel_work_proc_understanding}
Related work explores various approaches to understanding procedural actions, such as action recognition (AR), which focuses on classifying short video clips of procedure executions~\cite{Assembly101,MECCANO,EPIC-KITCHEN,ben2021ikea,xiong2020transferable}. 
Temporal action segmentation (TAS) aims to segment actions, given a long video of procedures~\cite{ding2023temporal,TAS,TAS2,romeo2025multi}. 
Other works aim to extract key moments and the sequence in which they occur in long-form videos ~\cite{lin2022learning,zhong2023learning,nagasinghe2024not,ashutosh2024video,shah2023steps,grauman2024ego,bansal2022my} \newtext{or aim to identify mistakes in procedural egocentric videos~\cite{prego}}.
\newtext{To this end, several approaches utilize narrations~\cite{zhong2023learning}, textual databases~\cite{lin2022learning}, and knowledge graphs~\cite{nagasinghe2024not} to enhance the spatio-temporal features with procedural knowledge.}
\newtext{Furthermore, several works explore various attention mechanisms for temporal modeling~\cite{shu2019hierarchical, tang2019coherence, yan2020higcin, shu2021spatiotemporal}.}
Each of the aforementioned tasks is \newtext{relevant and} of interest for procedure understanding, since it is particularly challenging to extract the spatio-temporal features required for acceptable performance. 
However, these tasks have a limited applicability \newtext{for use cases with industrial procedures, where a measure of completion and correctness is desirable or even required. 
For example, while approaches to identify procedural mistakes (\eg~\cite{prego}) can highlight potentially erroneous actions, it cannot guarantee the \emph{correct completion} of procedural steps: the absence of a correct step completion does not necessitate the presence of a procedural error, and thus can go unnoticed through such approaches. 
Similarly, action recognition and segmentation approaches provide only a classification of \emph{which} procedure step is being executed, without the notion of whether it has been done \emph{completely} and \emph{correctly}.}

While the aforementioned approaches emphasize human execution of procedural actions, assembly state detection (ASD) focuses on the transformation and progression of an object within a procedure, by detecting subsequent assembly states~\cite{ASD-1,ASD-2,ASD-3,ASDF,origami}.
The ASD task typically involves training an object detection network with bounding boxes and corresponding state labels, and performs well on datasets where images contain unobstructed views of assembly states.
While ASD can provide a measure of correctness for assembly states~\cite{schoonbeek2024supervised}, ASD approaches do not leverage temporal consistency on (even briefly) partially occluded objects.
Therefore, in egocentric videos, where assembly states are often partially occluded or out of frame, ASD models are unable to recognize object states.

Such occlusions are evident in procedural datasets such as Assembly101~\cite{Assembly101}, MECCANO~\cite{MECCANO}, Ikea ASM~\cite{ben2021ikea}, \newtext{HA4M~\cite{ha4m}}, and IndustReal~\cite{IndustReal}. \newtext{Although each dataset is similar in the task of assembling a relatively small object in industrial-like settings, there are notable differences between the datasets.}
To mitigate the impact of occlusions, Assembly101~\cite{Assembly101} and Ikea ASM~\cite{ben2021ikea} provide multiple viewpoints, while operators in IndustReal~\cite{IndustReal} were instructed to occasionally provide an unobstructed view of the object.
For MECCANO~\cite{MECCANO}, no such instructions were provided, resulting in a highly challenging dataset with only sparse non-occluded images of the toy motorcycle assembly.
Finally, only IndustReal provides CAD data to generate synthetic images of assembly states to supplement the real-world dataset.

\subsection{Representation learning with procedural data}
\label{ch2-subsec2:Representation learning for procedure learning}
Videos of procedure executions exhibit continuous movements, where objects change gradually, resulting in a high similarity among consecutive frames and larger variations between more distant frames. 
Consequently, representation learning has emerged as a valuable tool for understanding procedural video data, since it inherently learns frame embeddings that trace the procedure progression through latent space.
Bansal~\etal~\cite{bansal2022my} propose a self-supervised \textit{Correspond and Cut} framework to learn temporally corresponding features by aligning key steps in the embedding space across different videos of the same procedure.
Schoonbeek~\etal~\cite{schoonbeek2024supervised} propose a representation learning framework for assembly state recognition that enhances generalization to unseen states, demonstrating that supervised contrastive learning generates meaningful embedding spaces for procedural data.
\newtext{For skeleton-based action recognition in videos, contrastive learning has been used to learn multi-granularity anchor-contrastive representations in a semi-supervised manner~\cite{shu2022multi}.}
Finally, Vidalmata~\etal~\cite{temporal-embedding} propose a two-stage spatio-temporal approach for temporal action segmentation in instructional videos, employing a U-Net~\cite{ronneberger2015unet} to extract spatial features and an MLP for temporal embedding.
Inspired by these works, we propose a weakly supervised representation learning method to pre-train the spatial encoder within a spatio-temporal network.

\newtext{\section{Procedure step recognition}}
\label{sec:psr}
\newtext{This section formalizes the procedure step recognition (PSR) task as defined in~\cite{IndustReal}, outlines the evaluation metrics used to assess system performance, and reviews current limitations of state-of-the-art approaches based on assembly state detection.}

\subsection{Task definition}
The PSR task aims to detect procedure steps that are correctly completed by a person up to time $t$, given a sensory input $X_t = (x_1, x_2, \cdots x_t)$ (\eg, video) and $\mathcal{P} = \{p_0,\cdots p_N\}$, a descriptive set of the procedural actions to be performed~\cite{IndustReal}.
The predicted set of completed procedure steps $\hat{Y}_t$ consists of both the recognized steps and their respective completion times up to time $t$, formulated as

\begin{equation}
    \hat{Y}_t = \{(\hat{a}_{\sigma(0)},\hat{t}_{\sigma(0)}),\cdots (\hat{a}_{\sigma(m)},\hat{t}_{\sigma(m)})\},
\end{equation}

\noindent where $\hat{a}_{\sigma(i)}$ represents the predicted completion of action $a_i \in \mathcal{P}$ at time $\hat{t}_{\sigma(i)}$, and $m+1$ denotes the total number of recognized procedure steps. 
The ground truth procedure steps completed up to time $t$ are defined as
\begin{equation}
    Y_t = \{(a_{\rho(0)},t_{\rho(0)}),\cdots (a_{\rho(k)},t_{\rho(k)})\}, 
\end{equation}

\noindent where $a_{\rho(i)}$ is the predicted completion of the action $a_i \in \mathcal{P}$ at time $t_{\rho(i)}$, and $k+1$ denotes the total number of correctly executed steps.

\subsection{Evaluation metrics}
To quantify the performance of the PSR task, three evaluation metrics are used~\cite{IndustReal}. 
First, the procedure order similarity (POS) measures how closely the predicted step sequence aligns with the correct order using the edit distance~\cite{damerau:1964:damlevdistance}, and is defined as
\begin{equation}
    \text{POS} = 1 - \min({\frac{\text{DamLev}(Y,\hat{Y})}{|Y|},1)} \text{,}
\end{equation}
where the $\text{DamLev}(\cdot)$ is a weighted DamLev edit distance~\cite{DamLev} function. 
\newtext{Importantly, POS does not evaluate conformity to a canonical or ``correct'' step order. Instead, it measures how accurately the system recovers the order in which the user actually executed the procedure. This allows the POS metric to remain relevant even when steps are swapped or repeated, penalizing the model only if steps are recognized in an order different than the one executed by the user.}

The second metric, $F_1$ score, calculates the harmonic mean of recall and precision for all procedure steps.
A true positive (TP) is defined as the prediction of a procedure step $(\hat{a}_{\sigma(i)}, \hat{t}_{\sigma(i)})$ observed at or after the actual completion of the ground truth action $(a_{\rho(j)}, t_{\rho(j)})$, defined as

\begin{equation}
    (\hat{t}_{\sigma (j)} \geq t_{\rho (i)}) \wedge (a_i \in Y).
\end{equation}
 
A false positive (FP) is defined as a procedure step $\hat{a}_{\sigma(i)}$ that is detected prior to the actual completion of action $a_{\rho(j)}$, or if $a_i$ is not completed, defined as
 
 \begin{equation}
    (\hat{t}_{\sigma (j)} < t_{\rho (i)}) \vee (a_i \not \in Y).
 \end{equation}
 
 A false negative (FN) is defined as correctly completed procedure step $a_{\rho(j)}$ that is never recognized by the model, thus is defined as
 \begin{equation}
    (a_i \in Y) \wedge (a_i \not \in \hat{Y}).
 \end{equation}

Finally, the average delay ($\tau$) measured in seconds, quantifies the average time between the completion and corresponding recognition of a step and is defined as
 \begin{equation}
     \tau = \frac{1}{n} \sum_{i=0}^{n-1}(\hat{t}_{\sigma(i)} - t_{\rho(i)})\text{,}
 \end{equation}
 where $n$ is the number of TPs in the prediction $\hat{Y}$. 

Interpreting any of the three metrics in isolation can lead to a misleading assessment of system performance.
For example, a model that predicts all steps as completed only at the final frame of a video might achieve a high $F_1$ score while exhibiting a significantly large average delay ($\tau$) and POS scores.
Therefore, a comprehensive evaluation requires analyzing the three metrics together to obtain an accurate understanding of the model's overall performance.

\subsection{State-of-the-art and limitations}
The state-of-the-art approach to PSR relies exclusively on assembly state detections, indirectly achieving step recognition by inferring steps between consecutive assembly states using prior knowledge of the procedure~\cite{IndustReal}.
However, ASD can only accurately recognize object states that are minimally occluded and entirely within the image boundaries, since ASD algorithms predict the state of the whole assembly object requiring a comprehensive view. 
Consequently, with a single image offering only partial visibility, predictions would require assumptions about the state of non-visible parts.
This limitation causes significant delays between the completion and consequent recognition of steps, as unobstructed views of the object are rare, especially in egocentric videos.
Therefore, the proposed approach complements ASD predictions with a novel temporal network, optimized directly to recognize the completion of procedure steps, even under partial occlusions. 

%% file: 3_method.tex
\section{Method}
\label{sec:method}
\subsection{Overview}
This section presents the proposed approach to PSR, using the Spatio-Temporal Occlusion-Resilient Modeling for Procedure Step Recognition (STORM-PSR) framework.
STORM-PSR combines an assembly state detection (ASD) stream ($\mathcal{S}$) with a spatio-temporal stream ($\mathcal{T}$), as outlined in \cref{fig: teaser}.

\begin{figure}
    \centering
    \includegraphics[width=0.95\linewidth]{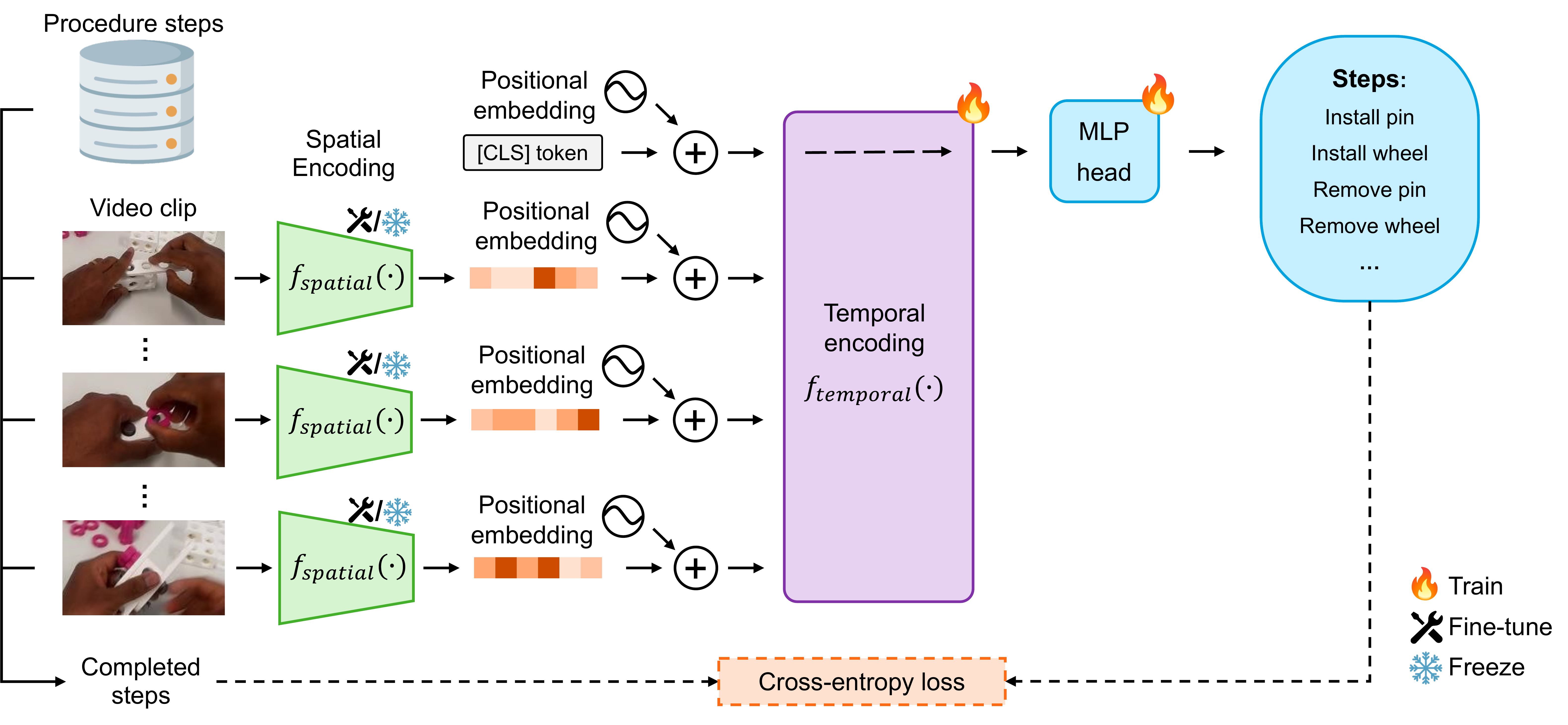}
    \caption{
    \newtext{
    Overview of the spatio-temporal stream of STORM-PSR and its training procedure.
    The frames of each video clip are passed through the spatial encoder $f_{\text{spatial}}(\cdot)$ to obtain per-frame embeddings. These embeddings are combined with a learnable [CLS] token, and augmented with 1-D positional encoding before being fed into the temporal encoder $f_{\text{temporal}}(\cdot)$. The final state of the [CLS] token provides a clip-level representation, which is fed into the MLP classification head that makes the final predictions. Because multiple procedural steps can be completed within a single clip, the MLP performs multi-label classification.
    During the training of the temporal encoder, the weights of $f_{\text{spatial}}(\cdot)$ can be either frozen or fine-tuned. The cross-entropy loss is used to optimize the model weights.
    }
    } 
    \label{fig: temporal_stream}
\end{figure}

Specifically, given an input video $X$ and a set of procedural actions $\mathcal{P}$, the objective of the two-stream method, STORM-PSR ($\mathcal{F}$), is to map $X$ and $\mathcal{P}$ to $\hat{Y}$, an ordered set of tuples containing each correctly completed procedure step $p\in \mathcal{P}$ and its moment of completion $t\in[0,T]$, where $T$ is the total duration of video $X$, such that

\begin{equation}
    \hat{Y} = \mathcal{F} (X, \mathcal{P}).
    \label{eq: input_output}
\end{equation}

\noindent Each stream produces a probability distribution over every procedure step $k\in \mathcal{P}$, denoted by $\hat{y}_{\mathcal{S},k}=\mathcal{S} (X, \mathcal{P})$ and $\hat{y}_{\mathcal{T},k}=\mathcal{T} (X, \mathcal{P})$ for the assembly state and spatio-temporal streams, respectively. 
The prediction probability $\hat{y}_k$ for the correct completion of step $k$ is then computed as the average with equal contribution from both streams,

\begin{equation}
    \hat{y}_k = 0.5 \cdot \hat{y}_{\mathcal{S},k} + 0.5 \cdot \hat{y}_{\mathcal{T},k}\,.
    \label{eq: averaging}
\end{equation}

As outlined in previous work, the ASD stream $\mathcal{S}$ performs well on video frames where the objects are entirely visible, but fails fundamentally when any parts are occluded or outside the image boundaries.
We propose to mitigate the issue of occlusion by complementing the ASD stream with a spatio-temporal stream, leveraging temporal consistency when parts are occluded.

\subsection{Supervised training of spatio-temporal model}
\label{method: temporal}
Frames showing full visibility of all assembly parts are rare, yet previous work relies entirely on such frames~\cite{IndustReal}. To avoid this, we propose directly optimizing a spatio-temporal model to recognize the correct completion of procedure steps.
By directly recognizing correctly completed procedure steps, only the parts relevant to each step need to be visible.
Additionally, the temporal model learns to leverage temporal continuity, enabling it to use information from components previously visible, even when temporarily occluded.

The overall structure and training methodology for the spatio-temporal model are illustrated in~\cref{fig: temporal_stream}.
The temporal model includes a pre-trained spatial encoder~$f_{\text{spatial}}(\cdot)$, a temporal encoder $f_{\text{temporal}}(\cdot)$, and a classification head.
The temporal encoder comprises a transformer architecture~\cite{transformer}.
The frames of a video clip $X_i$ are processed by the spatial encoder~$f_{\text{spatial}}(\cdot)$ and the first layer of it's projection head $g(\cdot)$ to obtain meaningful embeddings, following the approach in~\cite{chen2020big}.
Analogous to BERT~\cite{bert} and the vision transformer \cite{VIT}, these embeddings are then combined with learnable 1D~positional encodings~\cite{VIT}. 
A classification token, $[\text{CLS}]$, is added at the beginning of the spatial representation sequence to aggregate global information about the clip.
After propagating this input sequence through the temporal encoder, the output of the $[\text{CLS}]$ token is fed into the classification head.
The classification head is implemented using an MLP with one hidden layer, designed for multi-label classification at the clip level.
The output dimension of the classification head corresponds to the number of unique assembly components in the procedure $\mathcal{P}$, where each component is defined as the object part(s) necessary to complete a procedure step correctly. 

Similar to the video transformer network (VTN)~\cite{VTN}, the two encoders are trained separately for distinct objectives.
The spatial encoder is pre-trained in a weakly supervised manner to learn meaningful spatial features from similar assembly states, as outlined in the following section.
Subsequently, the temporal encoder is trained to capture temporal dynamics, by minimizing the multi-label binary cross-entropy loss $\mathcal{L}_{CE}$ to identify which components (if any) have undergone \textit{correctly} completed procedure steps in the video clip $X_i$.
The loss function $\mathcal{L}_{CE}$ is defined as 

\begin{equation}
    \mathcal{L}_{CE} = - \frac{1}{N} \sum_{i=1}^{N} \sum_{j=1}^{C} \left[ y_{ij} \log \hat{y}_{ij} + (1 - y_{ij}) \log (1 - \hat{y}_{ij}) \right] ,
    \label{eq: averaging}
\end{equation}

\noindent where $N$ is the number of samples in the batch, $C$ the number of steps to be completed, $y_{ij}\in\{0,1\}$ the binary ground-truth label for sample $i$ and class $j$, and $\hat{y}_{ij}$ the predicted probability.
A target value of zero is assigned to each procedure step that was not correctly completed within the clip, and a target value of one is assigned if it did.

To generate the step completion labels from video clips, an exclusive OR (XOR) operation is applied on the PSR labels between the first and last frame of the clip. 
This ensures that each clip is labeled based on the net step changes that occur within its duration.
For example, if two distinct steps are completed within a single video clip, the resulting label indicates the completion of both steps. 
However, if a removal step occurs first within the clip, followed by the re-installation of the same component, the label indicates no net change, since the component's state at the beginning and end of the clip remains the same.
This labeling methodology affects the training phase, as clips are sampled independently. During inference, however, video clips are processed consecutively, allowing the model to first recognize the completion of a removal step and later detect the subsequent installation step, once the removal action is no longer present in the sampled clip.

Similarly to~\cite{IndustReal}, a temporal filter is used to only consider predictions that are consistent across multiple frames. 
To this end, the confidence for each predicted step completion is accumulated over time until it reaches a threshold $T$, at which point the step completion and time-stamp are added to $\hat{Y}$.
On frames where no assembly state is recognized, the accumulated confidence scores for all incomplete steps decay at a fixed rate.

\subsection{Key-clip aware sampling}
\label{sec:method-sub:KCAS}


\begin{figure}
    \centering
    \includegraphics[width=1\linewidth]{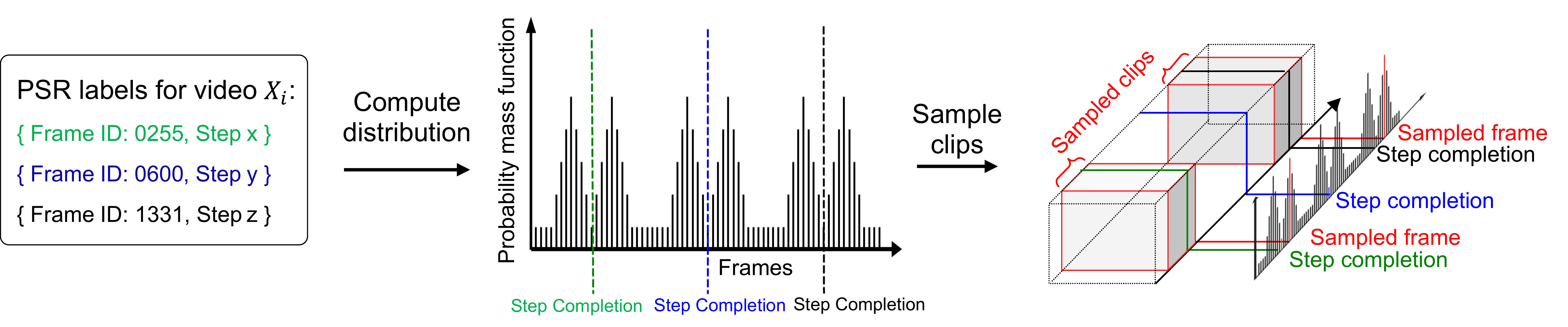}
    \caption{
    Principle view of generating the probability mass function with bimodal sampling used in key-clip aware sampling (KCAS). The KCAS over-samples hard-negative clips (directly before a step completion occurs) using the first Gaussian of the bimodal distribution, and positive clips (directly after the step completion) with the second Gaussian. Potentially ambiguous frames at the exact moment of step completions are sampled at a reduced rate, as well as frames where no steps are completed. \newtext{The long tails of the distribution deliberately suppress background clips far from any completion, preventing the mini-batch from being dominated by uninformative content.}
    }
    \label{fig: KCAS}
\end{figure}

Since most video clips do not contain step completions, uniformly sampling input clips for each mini-batch of the spatio-temporal model is inefficient.
To address this, we propose using key-clip aware sampling (KCAS), 
which over-samples video clips around the minority classes (step completions) compared to the background class (moments without step completions).
Specifically, KCAS creates a probability mass function, formed by two Gaussian distributions around each step completion time.
Let $T$ = $t_1, t_2, \cdots, t_m$ be the set of timestamps corresponding to the completions of procedure steps in video $X_i$, where $t_j$ represents the completion time of step $j$, and $\delta$ be the separation between the two Gaussian distributions for each step completion. The probability mass function $p_i(x)$ for sampling a clip \textit{ending at} frame $x$ is then defined as

\begin{equation}
    p_i(x) = \sum_{t_j \in T} \left[ g(x \mid t_j - \delta, \sigma) + g(x \mid t_j + \delta, \sigma) \right],
    \label{eq: kcas not norm}
\end{equation}

\noindent where $g(x \mid \mu, \sigma)$ is the probability mass function of a Gaussian distribution with mean $\mu$ and standard deviation $\sigma$. 
To ensure that $p_i(x)$ is a valid probability mass function, we normalize it by dividing $p_i(x)$ by the sum over all values of $x$.
When two PSR steps are completed at the same moment, they are seen as a single step completion for creating the sampling distributions, since the objective of KCAS is to over-sample clips where steps occur, regardless of the number of steps completed within this clip. 

A bimodal distribution is chosen over a single Gaussian because of potential ambiguity regarding the exact frame of completion for a procedure step, \eg the moment when a nut is tightened to the desired level.
We pose that those clips prior to completion can act as hard-negative samples, since they contain important information on the appearance of nearly completed steps.
Therefore, as demonstrated in~\cref{fig: KCAS}, KCAS explicitly samples \textit{fewer} video clips ending in the ambiguous moment, and \textit{more} clips that end prior to and after the step completions.
For each index sampled from the probability mass function, a temporal window of $w$~frames prior to this index is created, from which $N_w$~frames are sampled.
To reduce computational cost, a stride can be applied to limit the number of sampled frames ($N_w\leq w$)~\cite{simonyan2014two,wang2016temporal}.

\begin{figure}
    \centering
    \includegraphics[width=\linewidth]{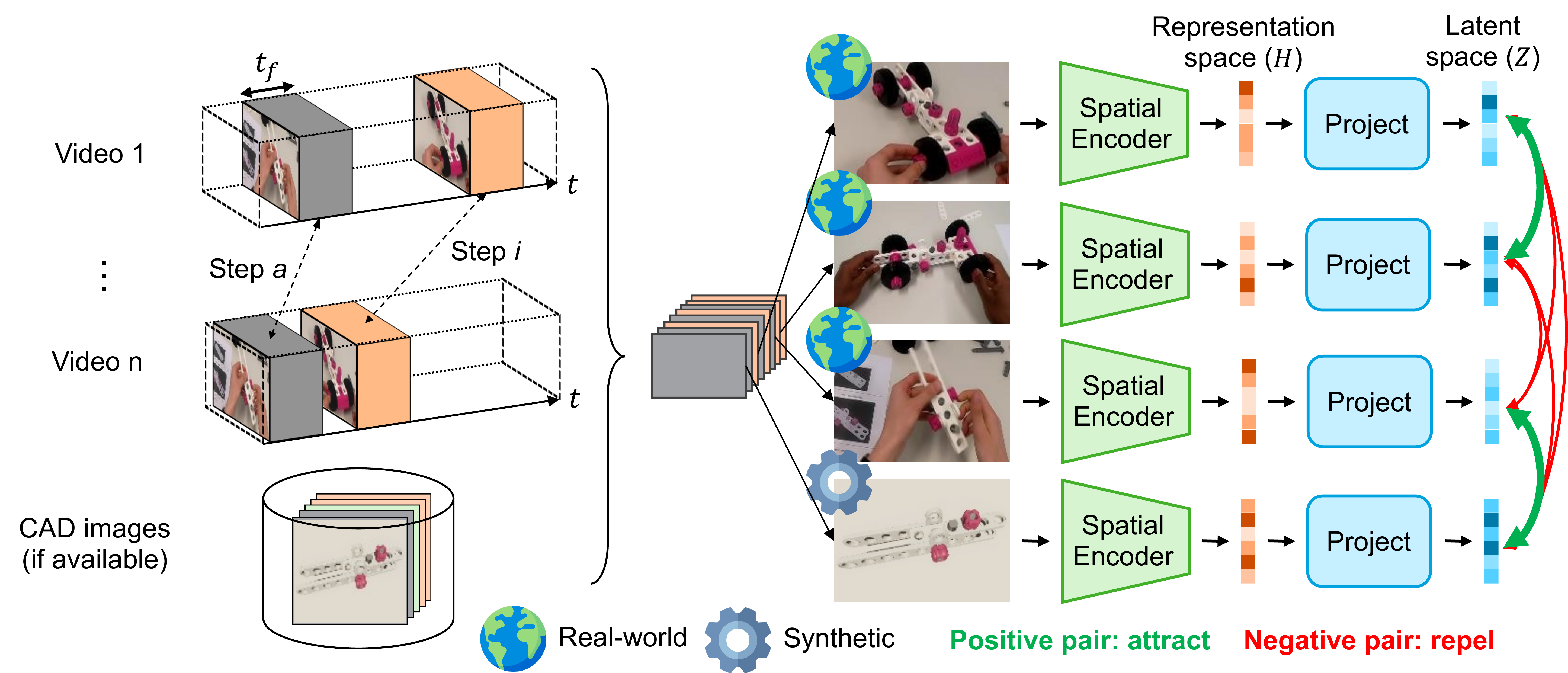}
    \caption{\newtext{Overview of the key-frame sampling (KFS)} weakly supervised pre-training. Each mini-batch consists of real-world images sampled around the time stamps of the PSR labels from different videos, with a sampling window $t_{\text{f}}$. When CAD data are available, synthetic images may be included. The sampled images are passed through a spatial encoder $f_{\text{spatial}}(\cdot)$, comprising a ViT-S~\cite{VIT} architecture, into representation space $H$, after which a three-layer MLP $g(\cdot)$ projects the embeddings into another space $Z$, on which the supervised contrastive loss~\cite{SupCL} is calculated.}
    \label{fig: KFA}
\end{figure}

\subsection{Weakly supervised training of spatial encoder}
\label{method: spatial}

\newtext{The goal of weakly supervised training of the spatial encoder is to encourage it to learn occlusion-robust state representations from limited supervision. Rather than requiring dense frame-level assembly state annotations, we leverage PSR labels as weak signals to group visual observations that correspond to the same procedural step, despite differences in viewpoint, occlusion, or tool presence. This enables robust feature learning from real-world observations, with synthetic data serving as an optional supplement that can enhance training.}

The spatial encoder $f_{\text{spatial}}(\cdot)$ is pre-trained with contrastive learning, outlined in detail in~\cref{fig: KFA}.
The approach, inspired by~\cite{schoonbeek2024supervised}, encourages $f_{\text{spatial}}(\cdot)$ to project instances of the same procedure steps, observed from different angles, videos, and occlusion levels, into similar embedding representations. 
Each image is processed by $f_{\text{spatial}}(\cdot)$, generating spatial feature embeddings $h$.
For $f_{\text{spatial}}(\cdot)$, an ImageNet-21K pre-trained ViT-S~\cite{VIT} model is employed, of which the last layer is replaced by a one-layer MLP to project the output into a 128-dimensional vector. 
The embeddings $h\in H$ are then projected into a new embedding space $z$ using a nonlinear projection head $g(\cdot)$, implemented as a three-layer multi-layer perceptron (MLP), as described in~\cite{chen2020big}.
The supervised contrastive loss~\cite{SupCL} ($\mathcal{L}_{\text{SupCon}}$) is employed to maximize the agreement of embeddings $z\in Z$ between different views and videos for identical assembly states, whilst minimizing the agreement of embeddings from different states, and is defined as

\begin{equation}
    \mathcal{L}_{\text{SupCon}} = \sum_{i\in I} - \log \frac{1}{\lvert P(i)\rvert} \sum_{p\in P(i)} \frac{\exp(\frac{g(f(x_i)) \cdot g(f(x_p))}{\tau})}{\sum \limits_{a\in A(i)} \exp(\frac{g(f(x_i)) \cdot g(f(x_a))}{\tau})},
    \label{eq:t2}
\end{equation}

\noindent where $A(i)$ the set of all indices in mini-batch $X$, and ${P(i)\equiv \{ p \in A(i) : (y_p = y_i)\}}$ is the set of indices of all positives, and $\tau$ is the temperature constant.

Mini-batches are created through key-frame sampling (KFS), leveraging PSR labels to gather images surrounding procedural step completions, as described in Algor.~\ref{algor:kfs}.
These PSR labels denote the actual completion of steps, regardless of the visibility of irrelevant parts of the assembled object within the frame. 
In contrast, assembly state detection annotations are present only when the entire object can be observed from a single image, requiring the entire object (including the irrelevant parts) to be largely unobstructed and within the frame.
Thus, pre-training on the PSR labels encourages similar assembly states to be grouped based on the relevant visible parts that constitute change.

KFS is weakly supervised, because it relies on sparse procedural step annotations rather than dense, per-frame supervision of assembly states. 
Specifically, the duration of visibility for the parts constituting a step completion is not annotated, so KFS must assume a fixed temporal window, denoted by $t_{\text{f}}$ and expressed in seconds, after each step completion.
As a result, the sampled frames are coarsely annotated, meaning there is no guarantee that they capture the actual parts undergoing change, except for the first frame ($t_{\text{f}}$=0s).
\newtext{This $t_{\text{f}}$ is the main hyperparameter of KFS. When the value is increased, more variability in occlusion levels and viewpoints are provided, potentially leading to richer feature embeddings. However, this may introduce redundant or noisy samples that do not constitute meaningfully to the step completion.}

Further hyperparameters of KFS are the number of real-world samples per step completion in each mini-batch ($N_{\text{sample}}$). \newtext{Additionally, if synthetic data (\eg~CAD models) are available, the hyperparameter that determines the number of generated images $N_\text{syn}$ can be set to readily incorporate the synthetic images in the spatial encoder training.}

\begin{algorithm}[t]
    \caption{Key-frame sampling}
    \label{algor:kfs}
    \DontPrintSemicolon 
    \KwIn{Video dataset $X$, PSR labels $Y$, temporal window per step $t_{\text{f}}$, \# real-world frames per state $N_{\text{sample}}$, \# assembly states $N_{\text{state}}$, \# synthetic images per state $N_{\text{syn}}$}
    \KwOut{Mini-batch with images $x$, labels $y$}
        
    $x$ $\gets$ empty list\\
    $y$ $\gets$ empty list\\
    $\text{FPS}$ $\gets$ 10 /*\textsl{Constant} */\\
    \For{\textsl{stateID} $\in$ $\{1\cdots N_{\text{state}}\}$}
    {
        \textsl{images}$\gets$ empty list\\
        \textsl{syn\_images}$\gets$ empty list\\
        \textsl{indices} $\gets$ indices from in $Y$ with \textsl{stateID}\\
        \For{\textsl{i} $\in$ \textsl{indices}}
        {
            append $X[\textsl{i} : \textsl{i}+t_{f}\cdot \text{
            FPS}]$ to \textsl{images}   
            append $X[\textsl{i} : \textsl{i}+N_{syn}]$ to \textsl{syn\_images}   
        }   
        \For{\textsl{i} $\in$  $\{0\cdots (N_{\text{sample}}+N_{\text{syn}})\}$}
        {   \eIf{i $<$ $N_{\text{state}}$}
            {
                $x[\textsl{i}]$ $\gets$ image from \textsl{images}\\
                $y[\textsl{i}]$ $\gets$ \textsl{stateID}\\
            }
            {
                $x[\textsl{i}]$ $\gets$ image from \textsl{syn\_images}\\
                $y[\textsl{i}]$ $\gets$ \textsl{stateID}\\
            }
        }  
    }
    \Return $x$, $y$
\end{algorithm}

\subsection{Assembly state detection stream}
As outlined in \cref{fig: teaser}, STORM-PSR combines the spatio-temporal stream ($\mathcal{T}$) with an assembly state detection stream ($\mathcal{S}$).
This ASD stream comprises the object detection-based PSR baseline from~\cite{IndustReal}, based on a YOLOV8-m backbone~\cite{YOLOv8}.
Specifically, this stream indirectly recognizes procedure steps $\hat{y}_{\mathcal{S}}$ in $X$ by determining which steps in $\mathcal{P}$ are required to transform the previously observed assembly state $s_{i-1}$ to the newly recognized state $s_i$.
While the spatio-temporal stream enhances robustness to occlusion, clean frames -- in which the entire object is visible -- still provide valuable and reliable assembly state information. 
By integrating both streams, STORM-PSR effectively leverages the strengths of each, using spatial cues when available and relying on temporal consistency when visibility is limited.

%% file: 4_experiment.tex
\section{Experiments}
\label{sec:exp}
In this section, the two-stream framework for procedure step recognition is implemented and evaluated on the IndustReal~\cite{IndustReal} and MECCANO~\cite{MECCANO} datasets.
Both datasets contain videos of construction-set toy assemblies recorded from an egocentric perspective.
A comparison on specifically these two datasets is interesting for three reasons.
First, operators in IndustReal are instructed to periodically provide unobstructed views of the toy car, which is not the case in MECCANO.
Therefore, MECCANO contains significantly more occluded frames, making it a more challenging dataset for PSR.
Second, the MECCANO dataset is 17\% longer (in hours) than IndustReal, but includes approximately half of the step completions and annotated ASD frames.
Finally, IndustReal offers synthetic data for all assembly states, providing additional data for the pre-training.
Hence, although the datasets are similar, STORM-PSR faces distinct challenges with each dataset.

\begin{figure}
    \centering
    \includegraphics[width=0.75\linewidth]{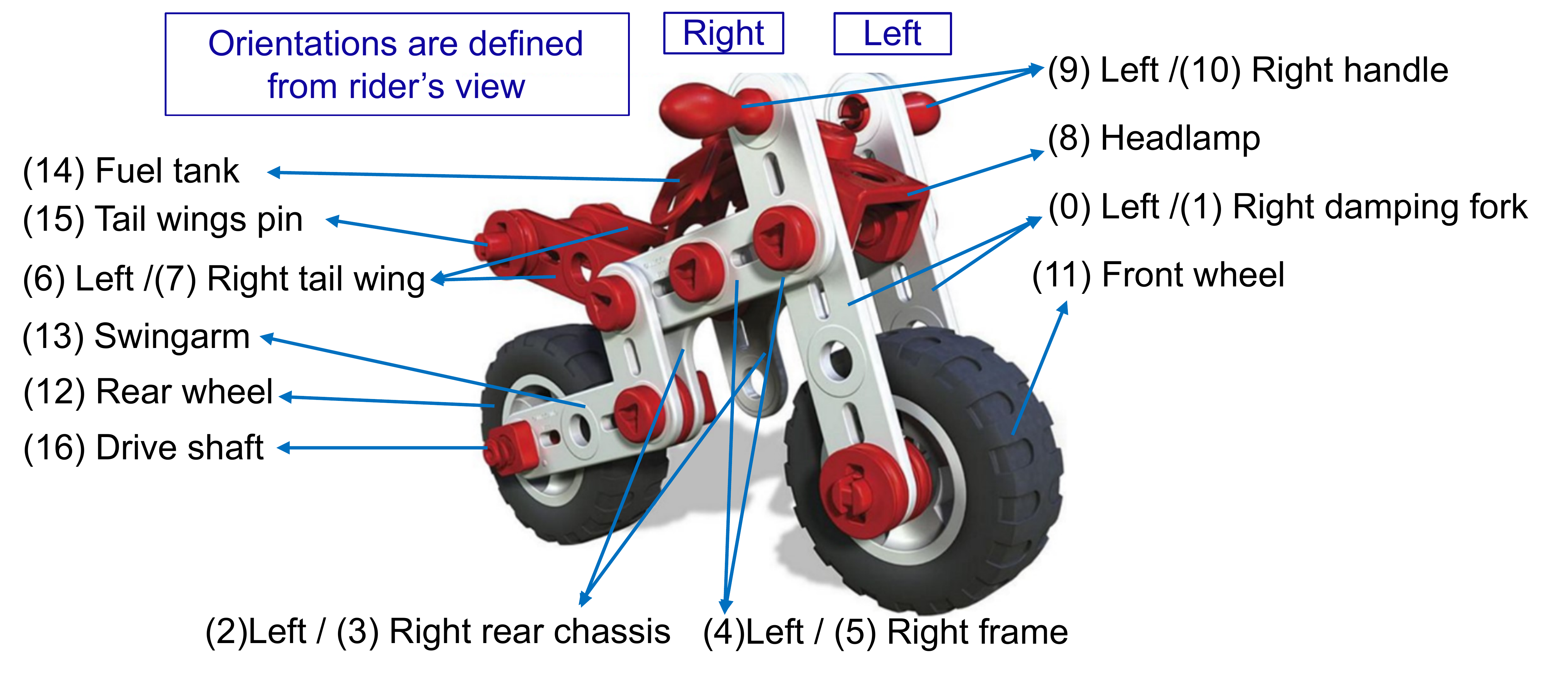}
    \caption{Definition of the 17 toy-motorcycle model assembly components in MECCANO.} 
    \label{fig: sup. components}
\end{figure}

\subsection{MECCANO annotation}
Because MECCANO does not include the required ASD and PSR annotations, we have manually annotated the dataset for both tasks, according to the same procedure as~\cite{IndustReal}.
These new labels are made publicly available to stimulate research on PSR.
For the MECCANO toy motorcycle, 11~assembly states consisting of 17~components, outlined in \cref{fig: sup. components}.

\begin{figure*}
    \centering
    \includegraphics[width=1\linewidth]{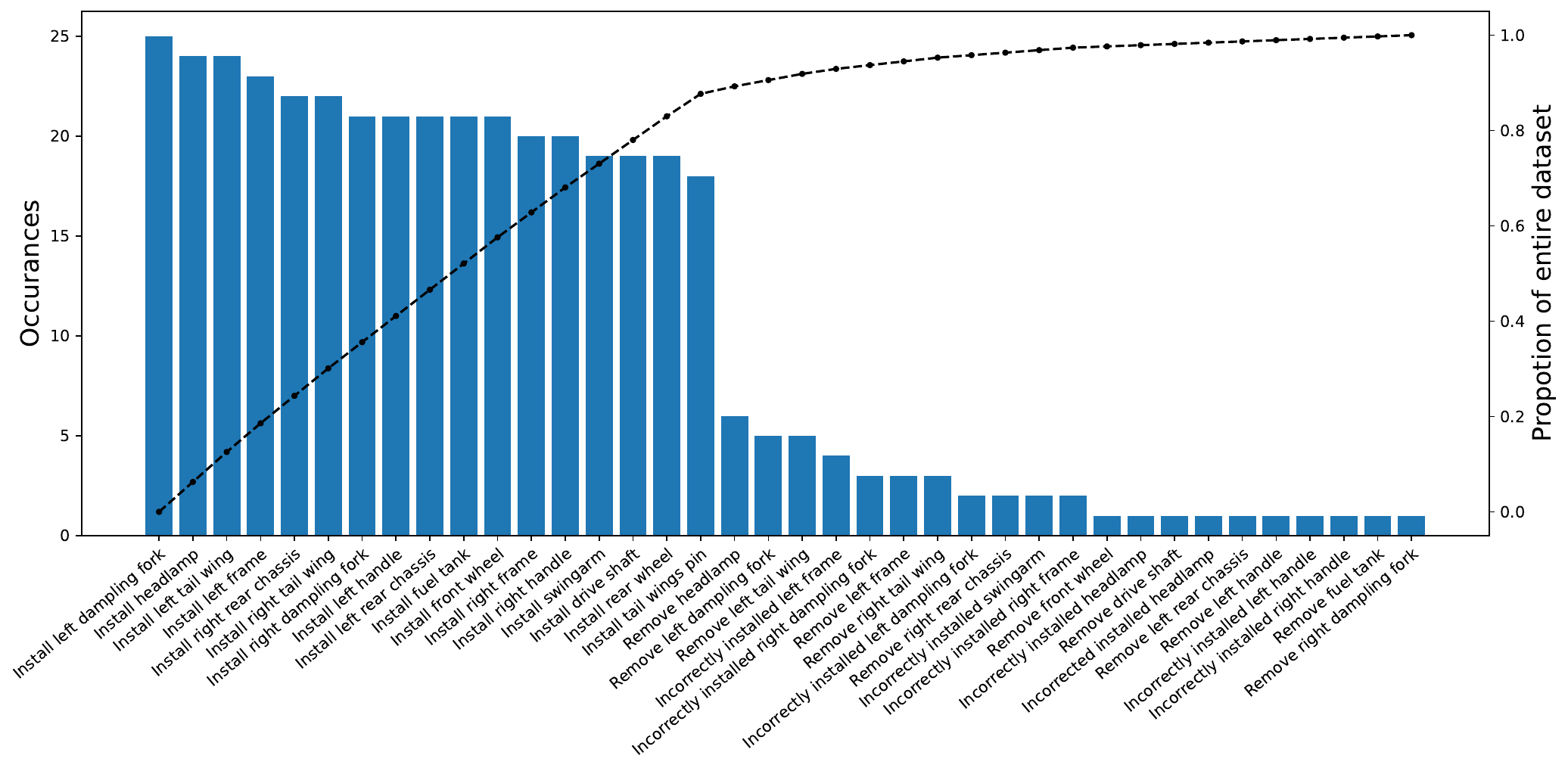}
    \caption{Distribution of the new procedure step recognition labels for the MECCANO dataset.} 
    \label{fig: sup. distribution}
\end{figure*}

To create the PSR labels, annotators were instructed to mark the assembly states and frame indexes when the participants complete an installation or removal action for each component. 
Figure \ref{fig: sup. distribution} demonstrates the installation, removal, and incorrectly installed class instances.
A long-tail distribution is observed, where the expected 17 correct installation actions (one per component) constitute approximately 85\% of the dataset.
The removal actions are scarce since they only occur after (infrequent) execution errors.
Based on the occurrences of the step completions, it is possible to see where mistakes are commonly made.
For instance, the action ``remove headlamp'' is commonly performed, indicating that a mistake is frequently made after ``install headlamp'' to correct it.
In total, the dataset contains 431~correctly executed procedure steps (25$\pm$4.0 steps per recording) and 30~incorrect step completions (1.5$\pm$0.9 per recording).

For the ASD annotations, the same procedure for annotation as for IndustReal is used.
Therefore, bounding boxes are only created if a human annotator can recognize the assembly state based on a single image, without using further temporal knowledge from the assembly videos. 
In total, 13.6K image frames are annotated with a bounding box, which is approximately half of the frames annotated for IndustReal (26.9K frames).
The training set contains 6,948~images, the validation set contains 3,053~images, and the test set comprises 3,656~images.
A benchmark performance for ASD on the MECCANO dataset is provided in \Cref{Appendix:MECCANO}.

\subsection{Implementation details}
\label{ch5.1:implementation details}
Both the spatial and the temporal encoder are optimized using SGD~\cite{SGD} with an initial learning rate of 10$^{-3}$, momentum of 0.9, five warm-up epochs, and a cosine-annealing learning rate scheduler.
All input images are resized to 224$\times$224 pixels.
The encoders are trained separately for 100~epochs, after which the spatial encoder is unlocked with learning rate 10$^{-5}$, to jointly converge further for 75~epochs.

For the spatial encoder, pre-trained with KFS, the number of frames eligible to be sampled after a step completion is empirically set to $t_{\text{f}}$=2 seconds and the number of states per batch is set to the total number of unique states in the dataset.
Per state, $N_{\text{sample}}$=16 images are sampled if no synthetic data are used, otherwise $N_{\text{sample}}$ and $N_\text{syn}$ are set to 8.
A margin of 0.01 and temperature of 0.07 are set for the SupCon loss, and the best epoch is selected based on the precision on the validation set.

For the temporal encoder, 6~stacked self-attention layers, 8~attention heads, and an MLP output size of~4,096 is used.
The batch size $B$ and temporal window size~$w$ are both set to~256.
Within this window, $N_w$=64 spatial embeddings are sampled with equal spacing.
For KCAS, the standard deviation $\sigma$ is set to 45~frames, and $\delta$, the separation between the two Gaussian distributions, is set to 80.
The best checkpoint is selected based on the $F_1$ score on the validation set.

As in~\cite{IndustReal}, predictions are accumulated over consecutive video frames until a cumulative confidence score of~$T$ is reached.
Confidence scores for prediction-less frames are decayed at a rate of~25\%.
For PSR inference with the two-stream approach, the thresholds $T$=6.0 for IndustReal and $T$=1.0 for MECCANO are established based on empirical results on the validation set.
\newtext{STORM-PSR operates at 75.1~frames/second on a NVIDIA A100 GPU.}

\input{final_result_keystep}

\subsection{Procedure step recognition results}
\label{ch5.5 PSR on two detaset}
\Cref{tab: final result} presents the performance of STORM-PSR across both datasets.
STORM-PSR reduces the average delay ($\tau$) by~26.1\% for IndustReal and 11.2\% for MECCANO, outperforming the state-of-the-art baseline.
Simultaneously, STORM-PSR achieves state-of-the-art performance on IndustReal for the POS and $F_1$ metrics.
On the MECCANO dataset, STORM-PSR improves the POS~(+6\%) but reduces the $F_1$ score~(-9\%).

The temporal stream alone does not outperform the object detection stream on either dataset.
This is likely because the datasets contain relatively few step completions, compared to the number of ASD-annotated frames. 
Additionally, spatio-temporal features are more challenging to learn due to increased dimensionality, the need to learn temporal coherence, and computational complexity~\cite{carreira2017quo,ji20123d,hara2018can}.
These factors lead to more false positives for the temporal stream compared to the ASD stream.
Combining the object-detection and temporal streams achieves state-of-the-art performance by leveraging the strengths of both approaches.

\begin{figure}
    \centering
    \begin{subfigure}[b]{0.75\linewidth}
    \centering
        \includegraphics[width=0.85\linewidth, trim={1cm 0.5cm 0.5cm 0}]{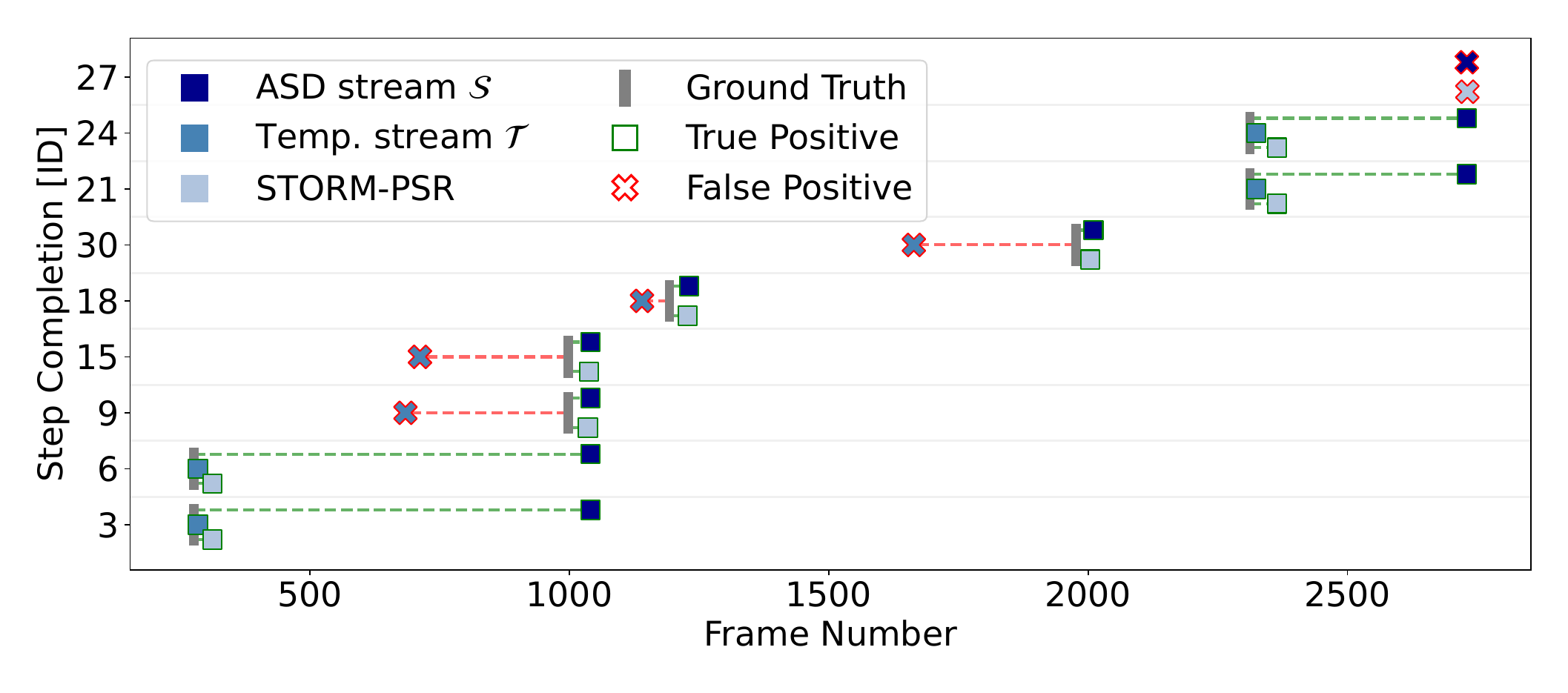}
        \caption{PSR predictions from the ASD and spatio-temporal streams, as well as STORM-PSR, which combines the both streams. \vspace{0.25cm}} 
        \label{exp:12_assy_3_4_simstepnet}
    \end{subfigure} 
    \hfill
    \begin{subfigure}[b]{0.75\linewidth}
        \includegraphics[width=1.0\linewidth]{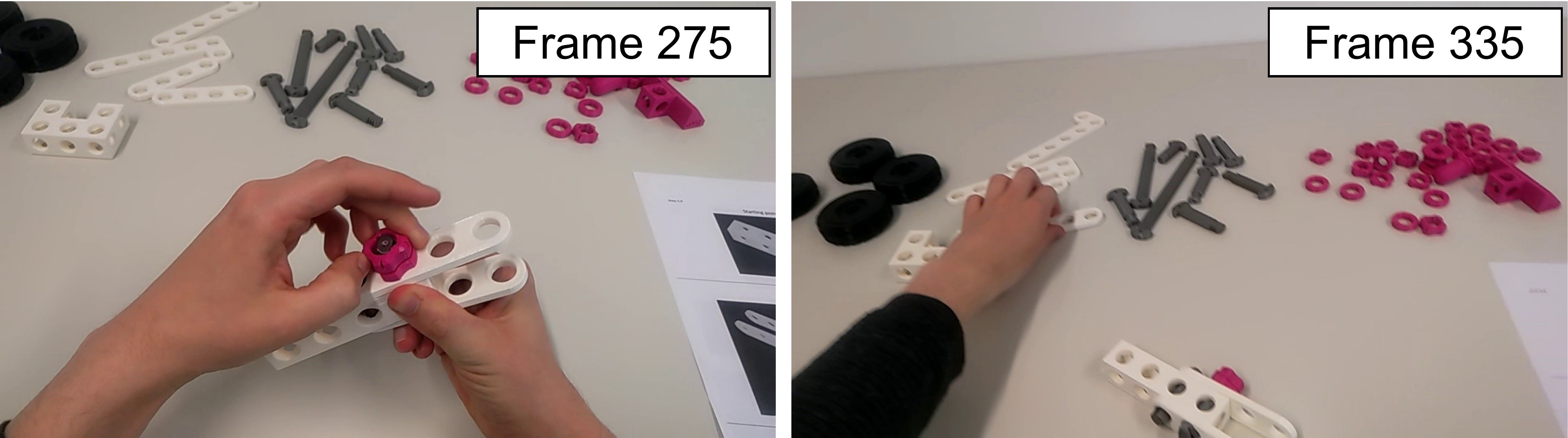}
        \caption{Two video frames demonstrating the completion of steps 3 and 6.} 
        \label{exp:12_assy_3_4_merge}
    \end{subfigure} 
 \caption{
   Qualitative demonstration of STORM-PSR on the IndustReal test set.
   The ASD stream fails to predict Steps~3 and 6, until it encounters unobstructed frames, resulting in a delay of 765~frames.
   The temporal stream recognizes Steps~3 and 6, but predicts false positives (with lower confidence) for other steps.} 
\label{fig: qual result on recording}
\end{figure}

A qualitative example of predictions from both individual streams and their combined output (STORM-PSR) is provided in~\cref{fig: qual result on recording}, along with images illustrating typical occlusions types.
This example clearly demonstrates how the two streams complement each other.
When the temporal stream predicts with high confidence, while the ASD stream predicts with lower confidence (\eg due to occlusions), procedure steps can still be recognized.
The STORM-PSR accumulates the confidences from both streams (reducing false positives), allowing steps to be recognized when either stream is highly confident, or both streams maintain moderate confidence over several frames.

\subsection{Assembly state detection on MECCANO}
The assembly state detection stream $\mathcal{S}$, using a YOLOv8-M model, is also evaluated for ASD on the MECCANO dataset~\cite{MECCANO}.
The model obtains a mean average precision (mAP@50) of only~0.120, compared to~0.838 for the best model on the IndustReal dataset~\cite{IndustReal}.
The latter is trained on real-world data, supplemented with images generated using the provided CAD models.
Nonetheless, even without leveraging these synthetic data, an mAP@50 of~0.753 is still achieved.
The performance gap between IndustReal and MECCANO clearly highlights the difficulty of recognizing entire assembly states in videos with frequent occlusions, since limited visibility of the object parts significantly impairs ASD models.

\subsection{Ablation studies}
\label{ch5.3: sampling strategy}

\begin{table*}
  \begin{center}
    {\small{
    \begin{tabular}{ l c c||c c c|c c c }
    \toprule
        {}                 & \multicolumn{2}{c||}{Sampling} & \multicolumn{3}{c|}{IndustReal~\cite{IndustReal}} & \multicolumn{3}{c}{MECCANO~\cite{MECCANO}}\\
        Spatio-temp. stream & KFS  & KCAS & POS $(\uparrow)$     & $F_1$ $(\uparrow)$     & $\tau$ [s] $(\downarrow)$ & POS $(\uparrow)$     & $F_1$ $(\uparrow)$     & $\tau$ [s] $(\downarrow)$\\
    \midrule
        LSTM       &    &     & 0.000            & 0.000            &   -             & 0.000            & 0.000            &  -  \\
        LSTM       &\cm &     & 0.180            & 0.303            & 33.2            & 0.047            & 0.146            & 222.8\\
        LSTM       &\cm & \cm & 0.204            & 0.365            & 40.9            & 0.037            &  0.248           & 121.6\\
        TCN        &    &     & 0.000            & 0.000            &   -             & 0.000            & 0.000            &   -  \\
        TCN        &\cm &     & 0.124            & 0.185            & 25.4            & 0.000            & 0.000            &   -  \\
        TCN        &\cm &\cm  & 0.195            & 0.414            & 49.4            & 0.134            & 0.181            & 153.7\\
        Transformer&    &     & 0.000            & 0.000            &   -             & 0.000            & 0.000            &   -  \\
        Transformer&\cm &     & 0.356            & 0.419            & 24.4            & 0.180            & 0.138            & 283.7\\
        Transformer&\cm & \cm & 0.497            & 0.506            & \textbf{14.2}   & 0.206            & 0.247            & 120.3\\
    \midrule
        STORM-PSR  &    &     & 0.467            & 0.511            & 62.6            & 0.269            & 0.358            & \underline{102.9}\\
        STORM-PSR  &\cm &     & \underline{0.766}& \underline{0.892}& 30.0            & \underline{0.329}& \underline{0.459}& 135.8\\
        STORM-PSR  &\cm &\cm  & \textbf{0.812}   & \textbf{0.901}   & \underline{15.5}& \textbf{0.377}   & \textbf{0.497}   & \textbf{88.6}\\
    \bottomrule
    \end{tabular}
    }}
\end{center}
\caption{\newtext{An overview of the impact that different backbones, KFS, and KCAS have on the performance of the temporal model. The evaluated backbones consist of a long short-term memory (LSTM), temporal convolutional network (TCN) and transformer  temporal model. Additionally, the performance of STORM-PSR (with the transformer backbone) is reported without KFS and KCAS, with only KFS, and with both. \textbf{Bold} text indicates the highest performance and \underline{underlined} text indicates the second best.}} 
\label{tab: ch2_ablation_model_kfs_kcas}
\end{table*}

\newtext{The following ablation studies isolate and quantify the contributions of key components within the proposed framework, including the sampling strategies (KFS and KCAS), temporal modeling backbones, sampling distributions, temporal receptive field, and key-frame sampling parameters. These experiments provide insight into design choices and their impact on procedure step recognition performance.}

\newtext{\subsubsection{Sampling techniques and different backbones for temporal modeling}}
\newtext{To evaluate the contributions of our sampling strategies (KFS and KCAS), as well as the the choice of temporal backbone, a three–way ablation is reported in \Cref{tab: ch2_ablation_model_kfs_kcas}. We compare long short-term memory networks (LSTM)~\cite{lstm}, temporal convolutional networks (TCN)~\cite{tcn}, and the transformer backbone. Each backbone is trained (i) without key-frame sampling (KFS) or key-clip aware sampling (KCAS), (ii) with KFS only, and (iii) with both KFS and KCAS. Finally, the end-to-end STORM-PSR results are reported for the same three settings to show how improvements in the temporal stream translate to the full two-stream system.}

\newtext{
The impact of key-frame sampling (KFS) and key-clip aware sampling~(KCAS) on the temporal stream performance is highlighted in~\Cref{tab: ch2_ablation_model_kfs_kcas}.
Without KFS pre-training of the spatial encoder, the temporal encoder is not able to learn meaningful spatio-temporal features and fails to make any PSR predictions.
With KFS pre-training, but without KCAS, the temporal encoder is able to learn from the meaningful spatial features.
However, a clear improvement in performance is noted when KCAS is used to effectively over-sample positive and hard-negative clips.
On both datasets, both the POS and $F_1$ improve 14\% to 79\% when KCAS is used.
}

\newtext{Among the evaluated temporal backbones, the transformer consistently outperforms the LSTM and TCN across all metrics when paired with both key-frame and key-clip aware sampling. On the IndustReal dataset, the transformer achieves the highest POS~(0.497), $F_1$~(0.506) score, and the lowest delay~(14.2s), with the LSTM and TCN backbones performing between~18\% and 250\%~worse. On MECCANO this trend is less pronounced due to the generally decreased scores on that dataset. These results indicate that the transformer is better suited for capturing long-range temporal dependencies required for recognizing procedure step completions under partial occlusions, compared to the LSTM and TCN.}

\subsubsection{Key-clip sampling for temporal modeling}
\label{ch5.3.1: stage 2}
A second ablation study evaluates the impact of the KCAS probability mass function by comparing uniform, Gaussian, and bi-modal sampling methods.
In this experiment, the same spatial encoder weights are used for all three sampling methods.
After training the temporal encoder, both encoders are jointly fine-tuned for 75~epochs.
As shown in~\Cref{tab: kca ablation}, the results confirm the hypothesis that sampling before and after key moments performs best, with the bi-modal distribution achieving the best performance across most metrics.
This outcome is in line with expectations, since the temporal encoder is exposed to more positive samples during training compared to uniform sampling.
In comparison to Gaussian sampling, KCAS selects fewer ambiguous clips by lowering the probability at the moment of step completions, while increasing hard-negative samples.
This encourages the temporal encoder to learn subtle yet critical differences that constitute step completions.

\input{kca_table}

\subsubsection{Temporal receptive field}
\label{ch5.4:ablation study on temporal receptive field}
\input{temporal_understanding}

The temporal encoder is tasked with recognizing (potentially occluded) steps based on spatial features within a given temporal window of size~$w$.
This final ablation study examines the effect of the size of the temporal window on the performance for the PSR task.
Furthermore, the study compares temporal self-attention using a transformer to temporal aggregation of spatial features with a simple MLP\newtext{, an LSTM, and an TCN.
These models are} trained with the same spatial encoder for different temporal window sizes $w$, with KCAS.
All frames within the window are sampled ($w$=$N_w$), except when $w$=256, where $N_w$=64 is obtained by applying a temporal stride of~4 to reduce the computational load.
Finally, the same spatial encoder is used for all experiments and is not fine-tuned together with the temporal encoder.

The results of this test, presented in~\cref{tab: temporal window}, demonstrate that increasing the temporal window size $w$ improves performance.
This is in line with expectations, since a larger temporal window offers more context to identify step completions.
Moreover, the attention-based temporal encoder outperforms the aggregation-based MLP with the same temporal window, although the difference is not substantial.

\newtext{\subsubsection{Time after step completions to sample for KFS}
The time after a step completion from which to sample real-world frames, denoted as $t_f$, is an important hyperparameter in the weakly-supervised key-frame sampling (KFS) strategy. 
It determines the diversity and quantity of visual representations available during contrastive pretraining of the spatial encoder.
A larger $t_f$ introduces more variability in occlusion levels, viewpoints, and context, potentially leading to richer and more robust feature embeddings. 
However, excessively increasing this value may introduce redundant or noisy samples that dilute the discriminative signal of step completions. 
This ablation study investigates how varying $t_f$ affects the downstream performance of the spatio-temporal stream and the overall STORM-PSR model, aiming to identify an optimal trade-off between training efficiency and feature quality.}

\begin{table}
  \begin{center}
    {\small{
    \begin{tabular}{ l| c c c| c c c }
    \toprule
    {$t_f$ [s]} & \multicolumn{3}{c|}{IndustReal~\cite{IndustReal}} & \multicolumn{3}{c}{MECCANO~\cite{MECCANO}}\\
    {} &  POS$\uparrow$&  $F_1\uparrow$& $\tau\downarrow$& POS$\uparrow$&  $F_1\uparrow$& $\tau\downarrow$\\
    \midrule
     0.5   & 0.308            & 0.511              & 43.3           & \underline{0.193}& \underline{0.241} & 192.8\\
     1.0   & 0.311            & \underline{0.514}  & 48.0           & 0.159            & 0.199             & \underline{114.7}\\
     2.0   & \textbf{0.406}   & \underline{0.514}  & \textbf{25.3}  & \textbf{0.351}   & 0.163             & \textbf{108.5} \\
     4.0   & \underline{0.350}& \textbf{0.526}     & \underline{39.8}& 0.127           & 0.233             & 252.2\\
     8.0   & 0.346            & 0.508              & 52.7           & 0.126            & \textbf{0.298}    & 150.4\\
    \bottomrule
    \end{tabular}
    }}
\end{center}
\caption{\newtext{Ablation study on the time after a step completion to sample from for the weakly-supervised training of the spatial encoder.}}
\label{tab: time_to_sample}
\end{table}

\newtext{\Cref{tab: time_to_sample} shows the results for this ablation study. The results show the intricate balance between the three metrics used to evaluate PSR. Nonetheless, for both datasets, a $t_f$ of 2 seconds results in the highest POS score and the lowest average delay $\tau$.
}

%% file: final_result_keystep.tex
\begin{table*}
  \begin{center}
    {\small{
    \begin{tabular}{ l| c c c| c c c }
    \toprule
    {Method} & \multicolumn{3}{c|}{IndustReal~\cite{IndustReal}} & \multicolumn{3}{c}{MECCANO~\cite{MECCANO}}\\
    {} &  POS$\uparrow$&  $F_1\uparrow$& $\tau\downarrow$& POS$\uparrow$&  $F_1\uparrow$& $\tau\downarrow$\\
    \midrule
    IndustReal~\cite{IndustReal} &  \underline{0.797}& \underline{0.891}& 21.0           & \underline{0.354} & \textbf{0.545} & \underline{99.8}  \\ 
    Spatio-temp. stream          &  0.497            & 0.506            & \textbf{14.2}  & 0.206             & 0.247          & 120.3 \\
    STORM-PSR                    &  \textbf{0.812}   & \textbf{0.901}   & \underline{15.5}& \textbf{0.377}   & \underline{0.497}& \textbf{88.6}  \\ 
    \bottomrule
    \end{tabular}
    }}
\end{center}
\caption{\newtext{Performance of STORM-PSR and its two streams for procedure step recognition on the IndustReal~\cite{IndustReal} and MECCANO~\cite{MECCANO} datasets. \textbf{Bold} text indicates the highest performance and \underline{underlined} text indicates the second best.}}
\label{tab: final result}
\end{table*}

%% file: kca_table.tex
\begin{table}
  \begin{center}
    {\small{
    \begin{tabular}{ l| c c c| c c c }
    \toprule
    {Sampling} & \multicolumn{3}{c|}{IndustReal~\cite{IndustReal}} & \multicolumn{3}{c}{MECCANO~\cite{MECCANO}}\\
    {} &  POS$\uparrow$&  $F_1\uparrow$& $\tau\downarrow$& POS$\uparrow$&  $F_1\uparrow$& $\tau\downarrow$\\
    \midrule
    Uniform & 0.356             & \underline{0.419}& 24.4             &  0.180            &  0.138            & 283.7\\ 
    Gaussian & \underline{0.419}& 0.382            & \underline{22.4} &  \textbf{0.212}   &  \underline{0.175}& \underline{143.7} \\
    \newtext{KCAS} & \textbf{0.497}    & \textbf{0.506}   & \textbf{14.2}    & \underline{0.206}&  \textbf{0.247}   & \textbf{120.3} \\ 
    \bottomrule
    \end{tabular}
    }}
\end{center}
\caption{\newtext{Comparison of two baseline sampling approaches and the novel key-clip aware sampling (KCAS). For uniform sampling, clips are randomly selected, regardless of the corresponding class label. For Gaussian sampling, all clips with at least one step completion are regarded as positive samples, which are over-sampled accordingly. KCAS uses a bimodal sampling distribution, to explicitly over-sample (i) hard negative samples directly before step completions and (ii) positive samples, while under-sampling the negatives.}}
\label{tab: kca ablation}
\end{table}

%% file: temporal_understanding.tex
\begin{table}
    \begin{center}
        {
        \small{
        \begin{tabular}{ l| c c c| c c c }
        \toprule
        {Model}   & \multicolumn{3}{c|}{IndustReal~\cite{IndustReal}} & \multicolumn{3}{c}{MECCANO~\cite{MECCANO}}\\
                 & POS$\uparrow$        & $F_1\uparrow$          & $\tau\downarrow$     & POS$\uparrow$        & $F_1\uparrow$        & $\tau\downarrow$\\
        \midrule
        MLP-16    & 0.226               & 0.346                  & 42.2                 & 0.168                & 0.166               & 162.7\\ 
        MLP-64    & 0.285               & 0.357                  & 34.0                 & 0.202                & 0.130               & 172.3\\ 
        MLP-256   & \textbf{0.407}      & 0.330                  & \textbf{23.0}        & 0.249                & 0.156               & \underline{115.5}\\ 
        \newtext{LSTM-16}   & 0.121               & 0.238                  & 27.2                 & 0.190                & 0.189               & 194.5\\
        \newtext{LSTM-64}   & 0.200               & 0.286                  & 30.2                 & 0.173                & 0.207               & 258.9\\
        \newtext{LSTM-256}  & 0.277               & 0.470                  & 30.3                 & 0.156                & \textbf{0.225}      & 132.4\\
        \newtext{TCN-16}    & 0.119               & 0.144                  & 34.2                 & 0.172                & 0.113               & 163.9\\
        \newtext{TCN-64}    & 0.203               & 0.235                  & 25.6                 & 0.185                & 0.130               & 186.1\\
        \newtext{TCN-256}   & 0.265               & \underline{0.502}      & 43.1                 & 0.195                & 0.151               & 127.8\\
        Tr.-16    & 0.228               & 0.218                  & 38.1                 & 0.207                & \underline{0.212}   & 125.7\\ 
        Tr.-64    & 0.304               & 0.364                  & 45.6                 & \underline{0.289}    & \underline{0.212}   & 137.0\\ 
        Tr.-256   & \underline{0.406}   & \textbf{0.514}         & \underline{25.3}     & \textbf{0.351}       & 0.163               & \textbf{108.5}\\
        \bottomrule
        \end{tabular}
        }
        }
    \end{center}
\caption{\newtext{Comparison of PSR performance of a multi-layer perceptron (MLP), long short-term memory (LSTM), temporal convolutional network (TCN) and transformer (Tr.) temporal model. Each model is trained with three receptive fields, namely 16, 64, and 256 frames, which is indicated behind the model abbreviation (\ie Tr-16 is the transformer model with a window size of 16 frames).}}
\label{tab: temporal window}
\end{table}

%% file: 5_discussion.tex
\section{Discussion}
\label{sec:discussion}
This work aims to reduce delays and enhance robustness of procedure step recognition algorithms by enabling step predictions even when parts of the assembly are occluded.
This is achieved using a two-stream framework called STORM-PSR, where one stream focuses on robust state detection under minimal occlusions, while the second directly predicts steps based on spatio-temporal features.
STORM-PSR is evaluated on two datasets, IndustReal~\cite{IndustReal} and MECCANO~\cite{MECCANO}.
For the latter, PSR and ASD labels were initially unavailable, so these were manually annotated, and are published alongside this work.
On IndustReal~\cite{IndustReal}, STORM-PSR achieves state-of-the-art performance, and on MECCANO it establishes a new benchmark.

STORM-PSR significantly reduces the delay in recognizing the completion of steps.
While the delay reduction is in line with expectations, a larger improvement for the temporal stream of STORM-PSR on the MECCANO dataset was expected due to its high number of occluded step completions.
In fact, a 26\%~delay reduction ($\tau$) is observed on IndustReal, compared to only~11\% on MECCANO.
\newtext{Simultaneously, STORM-PSR improves the performance on both other metrics ($F_1$-score and POS) compared to the baseline.
On the other hand, STORM-PSR on the MECCANO dataset only improves the POS metric, while reducing the $F_1$-score by approximately 0.05.
On this point, the interaction between the three metrics is important to highlight, because the average delay $\tau$ can only be calculated on true positive recognitions of the model~\cite{IndustReal}.
Therefore, while the benefit of STORM-PSR on IndustReal is readily observed (since all metrics improve), the benefit on MECCANO is less pronounced and more complex to understand.
First, spatio-temporal models generally require a larger amount of data compared to models relying exclusively on spatial features, without the complexity of temporal learning~\cite{hara2018can,carreira2017quo}.
The MECCANO and IndustReal are of approximately the same total video length, but MECCANO contains only 1.1~procedure steps are completed per minute of video data, compared to 2.2~steps per minute in IndustReal.
We hypothesize that this higher data sparsity in MECCANO poses challenges for spatio-temporal learning, disproportionally affecting the spatio-temporal stream (and thereby STORM-PSR) compared to the ASD stream, that only relies on spatial features. 
This difference is further amplified by the lack of synthetic data for MECCANO, which is used in STORM-PSR on IndustReal during pre-training of the spatial encoder.
Second, the reduction in delay obtained by MECCANO may occasionally lead to earlier predictions that slightly precede true step completions, counting them as false positive predictions and thus negatively impacting $F_1$-score.
Finally, the frequency and nature of occlusions in the MECCANO dataset differs from that in the IndustReal dataset on three important aspects.
Firstly, operators in IndustReal, contrary to MECCANO, are asked to provide unobstructed views with some frequency, resulting in more frequent images without occlusions.
Secondly, MECCANO participants use a tool, whereas the participants in the IndustReal dataset only use their hands, creating a different type of occlusion.
Thirdly, the IndustReal dataset is recorded with a projected virtual box that outlines the FoV of the cameras. 
Therefore, participants at least somewhat consistently keep the object of interest within this FoV.
Contrarily, videos in the MECCANO dataset frequently do not contain any frames with the object visible for the entire length of the temporal window, limiting the performance of STORM-PSR on this data.
}

To address this data sparsity in the MECCANO dataset without additional video recordings and labor-intensive annotations, MECCANO could be further extended with synthetic data. \newtext{Given the commercial application of the MECCANO parts, using CAD models for creating synthetic data might not be feasible. Instead, approaches such as neural radiance fields (NeRFs)~\cite{mildenhall2021nerf} may allow for synthetic view generation based on real-world captures, providing an alternative to CAD-based synthesis.}
Additionally, video augmentations can be used to increase data variability~\cite{kwon2020rotationally,cauli2022survey}.

Further experiments may explore larger temporal windows, as the largest tested in this study spans 256~frames, corresponding to video clips of 25.6~seconds in IndustReal and 21.3~seconds in MECCANO.
Considering the time between step completions in both datasets, increasing this window size may be beneficial.
To use larger temporal windows without encountering computational limitations, frames can be sampled at a reduced rate.
Alternatively, a selection procedure can be applied to retain only the most informative frames \newtext(\eg through attention mechanisms), since not all frames are equally valuable~\cite{wang2021adaptive,han2023dynamic}, \newtext{and approaches that consider the entire video sequence can be explored~\cite{donahue2015long}}.
\newtext{The computation cost of STORM-PSR compared to the baseline, consisting only of the ASD stream, is significantly higher.
Specifically, the ASD stream operates at 284.8 FPS on an A100 GPU, while STORM-PSR with the transformer backbone operates at 75.1 FPS.
With the LSTM backbone, STORM-PSR achieves 72.9 FPS, and with a TCN backbone it operates at 75.2 FPS.
This trade-off between computational complexity and improved recognition delay is an important factor for practical applicability. Although STORM-PSR requires more computation, its ability to substantially reduce recognition delay provides a clear advantage in scenarios where timely predictions are essential.
}

\newtext{Another interesting direction of future research is regarding the fusion of the two streams. In the present work, a linear late-fusion approach is adopted to reduce the risk of overfitting, particularly given the limited size of the two evaluated datasets. Introducing learning-based fusion techniques can allow the model to weight the two streams differently to optimally exploit the complementary strengths of either stream. Especially in the context of larger-scale data, such adaptive fusion mechanisms may further improve performance. Another promising direction for future work lies in adapting existing methods originally developed for related tasks to the PSR setting. While these methods were not designed for PSR and require significant adjustments, they can offer comparative baselines and help contextualize the strengths and limitations of STORM-PSR.}

%% file: 6_conclusion.tex
\section{Conclusion}
\label{sec:conclusion}
This work proposes STORM-PSR, a novel approach to procedure step recognition based on spatio-temporal features. 
The STORM-PSR framework combines assembly state detections, which are robust when objects are entirely visible and unoccluded, with a temporal model to predict step completions.
This is the first work to directly optimize for PSR, rather than inferring step completions from observations of entire assembly states.
To effectively train the temporal model, we introduce key-frame and key-clip aware sampling strategies. 
STORM-PSR is evaluated on an existing PSR dataset, obtaining state-of-the-art performance, and on a new dataset where it sets a benchmark.

The results demonstrate that waiting for unobstructed views of entire assembly states leads to significant delays between the completion and corresponding recognition of procedure steps.
We demonstrate that this delay can be reduced using a spatio-temporal model capable of learning from visible assembly parts, without requiring assumptions about occluded parts.
A limitation of the proposed method is that the model does not explicitly learn the entire procedure due to its restricted temporal receptive field. 
Furthermore, to learn the spatio-temporal dynamics, the model requires significant training examples, which are cumbersome to collect and annotate.
In future work, this data scarcity may be mitigated by, \eg using a video-frame selection procedure that retains only the most informative frames for processing by the temporal network.

\section*{Acknowledgements}
This work is partially executed ASML Research and received funding from ASML and TKI grant number TKI2112P07.

%% file: X_suppl.tex
\appendix

\section{Assembly state detection on MECCANO}
\label{Appendix:MECCANO}

\begin{table}[!b]
\caption{Definition of assembly state with corresponding assembly state label and components being installed compare to the previous assembly state.}
\label{tab: sup meccano states}
\begin{tabular}{ c| l | c}
\toprule
Assembly state & \multicolumn{1}{c|}{Installed components}  & \multicolumn{1}{c}{Assembly state label} \\ \midrule
0              & Initial state                           & 00000000000000000             \\
1              & left damping fork, left frame, headlamp & 10001000100000000             \\
2              & right damping fork, right frame         & 11001100100000000             \\
3              & left handle, right handle               & 11001100111000000             \\
4              & left rear chassis, left tail wings      & 11101110111000000             \\
5              & right rear chassis, swimarm             & 11111110111001000             \\
6              & right tail wing                         & 11111111111001000             \\
7              & driving shaft                           & 11111111111001001             \\
8              & fuel tank                               & 11111111111001101             \\
9              & front wheel                             & 11111111111101101             \\
10             & rear wheel                              & 11111111111111101             \\
11             & tail wing pin                          & 11111111111111111             \\ \bottomrule
\end{tabular}
\end{table}

\begin{figure}
     \centering
     \begin{subfigure}{0.75\linewidth}
     \includegraphics[width=1\linewidth]{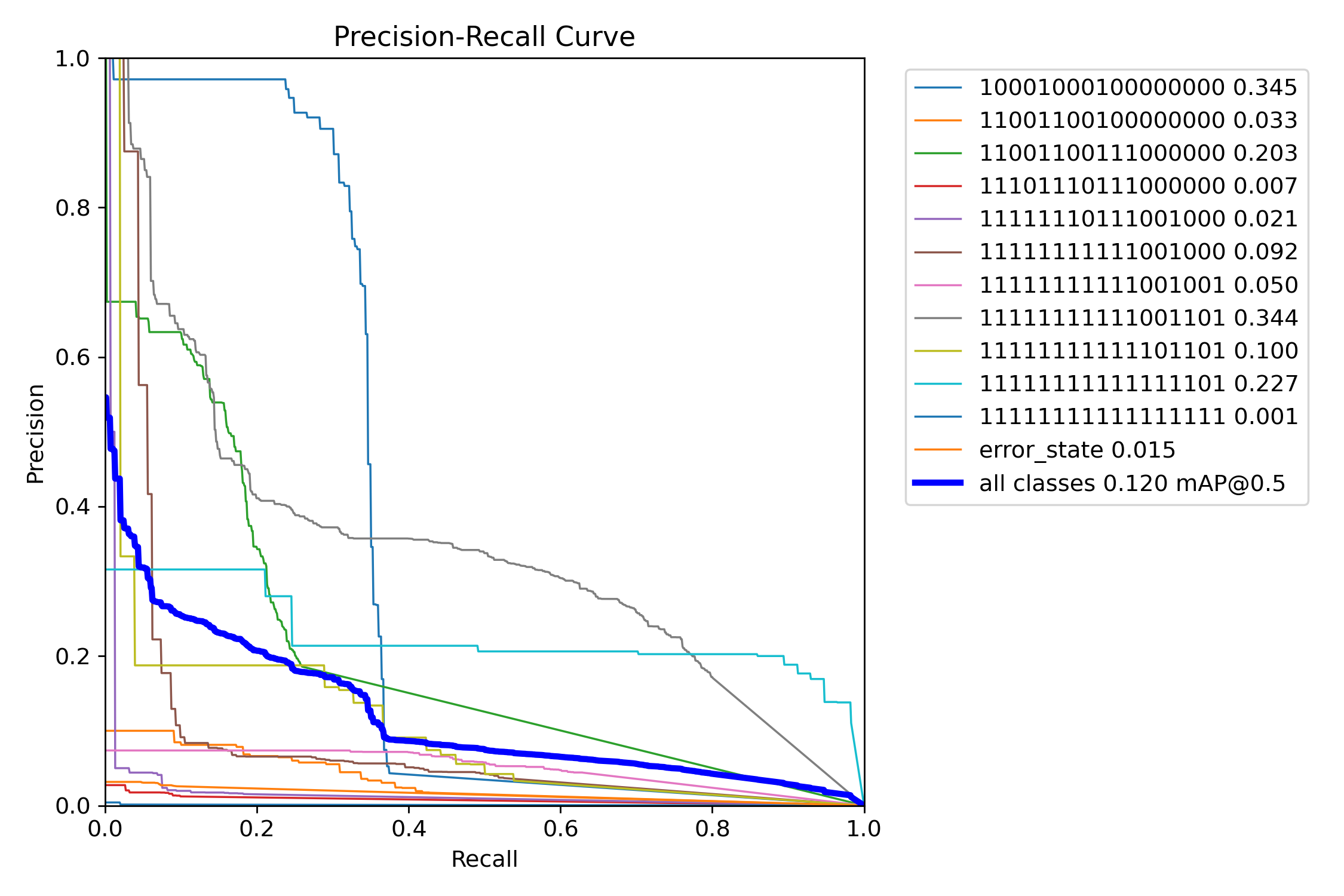}
     \caption{Precision-recall curve.} 
     \label{fig: asd pr curve}
    \end{subfigure} 
    \begin{subfigure}{0.9\linewidth}
     \includegraphics[width=1\linewidth]{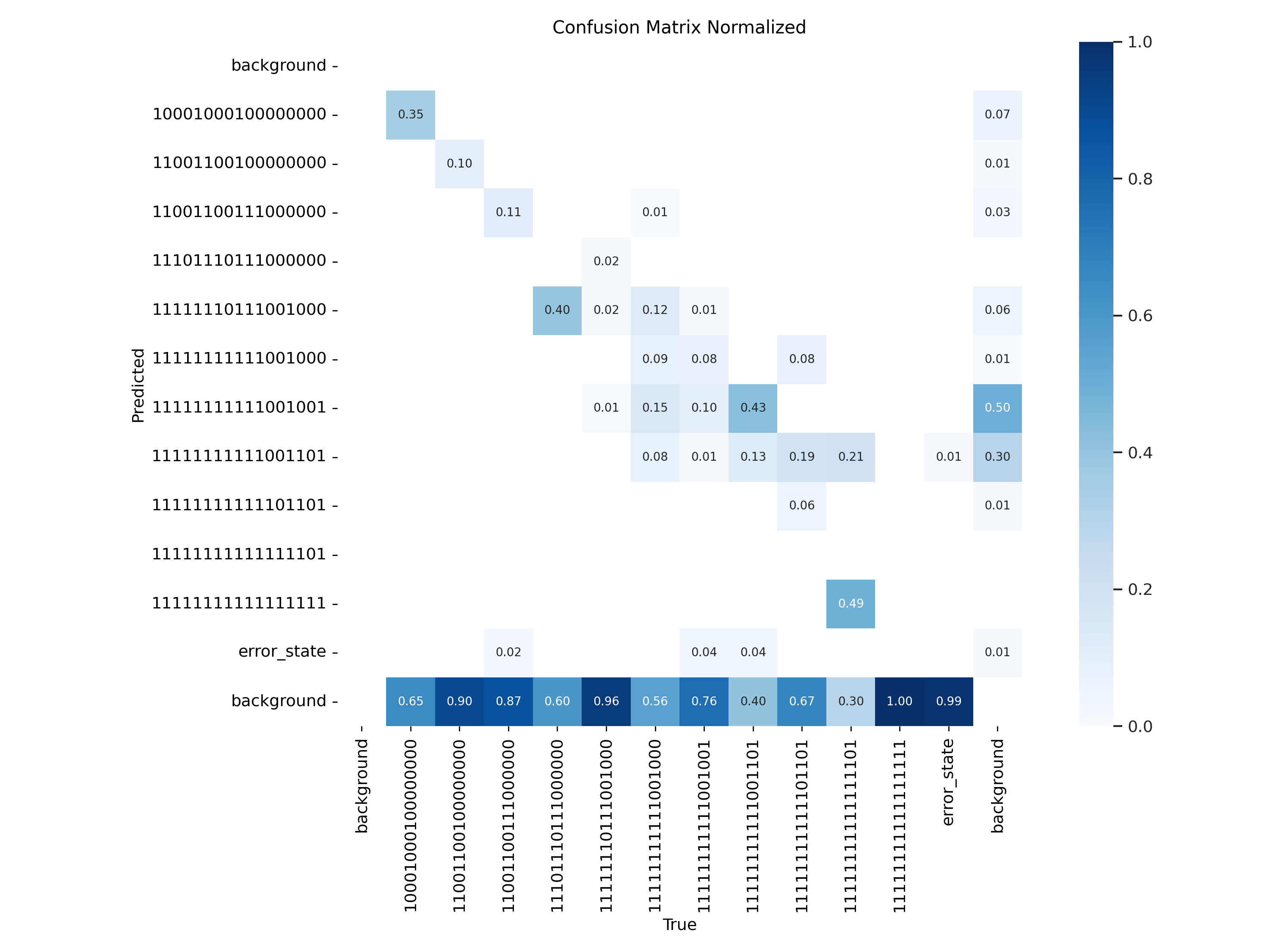}
     \caption{Normalized confusion matrix.} 
     \label{fig: asd confusion}
    \end{subfigure} 
 \caption{Assembly state detection performance on MECCANO.} 
\label{fig: meccano asd}
\end{figure}

The assembly states for assembly state detection are outlined in~\cref{tab: sup meccano states}, where each component corresponds to the components highlighted in \cref{fig: sup. components} of the main paper. 

The MECCANO test-set performance of a YOLOv8-M model, trained for ASD on the newly annotated labels, is shown in~\cref{fig: meccano asd}.
These results clearly outline a performance difference with IndustReal.
The model, trained for ASD on MECCANO, obtains a mean average precision (mAP@50) of 0.120, compared to 0.838 for the best model, trained on real-world and synthetic data, on the IndustReal dataset.
The best model of IndustReal that does not use synthetic data, still achieves an mAP@50 of 0.753.
This performance gap demonstrates the difficulty of recognizing entire assembly states on the MECCANO dataset and highlights the need for alternative approaches, such as STORM-PSR.